\documentclass[11pt]{article}

\usepackage[preprint]{acl}
\usepackage{times}
\usepackage{latexsym}
\usepackage[T1]{fontenc}
\usepackage[utf8]{inputenc}
\usepackage{microtype}
\usepackage{inconsolata}
\usepackage{graphicx}
\usepackage{booktabs}
\usepackage{pifont}
\usepackage{tikz}
\usetikzlibrary{shadows}

\definecolor{wordblue}{RGB}{32,84,159}
\newcommand{\docframe}[2]{%
\begin{tikzpicture}
    \node[inner sep=0pt] (img) {\includegraphics[width=0.95\linewidth]{#1}};
    \shade[left color=black!8, right color=white]
        (img.south east) rectangle ([xshift=3pt, yshift=12pt]img.north east);
    \shade[top color=black!8, bottom color=white]
        (img.south west) rectangle ([yshift=-3pt]img.south east);
    \fill[white] (img.south west) rectangle ([yshift=12pt]img.north east);
    \node[inner sep=0pt] (img2) {\includegraphics[width=0.95\linewidth]{#1}};
    \draw[gray!40, line width=0.3pt] (img2.south west) rectangle ([yshift=12pt]img2.north east);
    \fill[wordblue] (img2.north west) rectangle ([yshift=12pt]img2.north east);
    \node[anchor=west, font=\scriptsize\sffamily\bfseries, text=white]
        at ([yshift=6pt, xshift=4pt]img2.north west) {#2};
\end{tikzpicture}}
\newcommand{\cmark}{{\color{green!70!black}\ding{51}}}
\newcommand{\xmark}{{\color{red}\ding{55}}}
\newcommand{\ours}{WordInferBench}

\title{\ours{}: Can Agents Infer How to Edit from Word Document Context?}
\author{
  Jiayu Ding\textsuperscript{1*} \quad
  Yupan Huang\textsuperscript{2} \quad
  Jinyoung Park\textsuperscript{3*} \quad
  Yilin Jia\textsuperscript{2} \quad
  Wenshan Wu\textsuperscript{2} \\
  \bfseries Tengchao Lv\textsuperscript{2} \quad
  Shaohan Huang\textsuperscript{2} \quad
  Yuzhong Zhao\textsuperscript{2} \quad
  Xiangyang Zhou\textsuperscript{2} \quad
  Tao Ge\textsuperscript{2} \\
  \bfseries Sun Mao\textsuperscript{2} \quad
  Xun Wang\textsuperscript{2} \quad
  Huitian Jiao\textsuperscript{2} \quad
  Si-Qing Chen\textsuperscript{2} \quad
  Lei Cui\textsuperscript{2} \\
  \bfseries Nanning Zheng\textsuperscript{1$\ddagger$} \quad
  Furu Wei\textsuperscript{2$\ddagger$} \\[0.5em]
  \normalfont
  \textsuperscript{1}Xi'an Jiaotong University \quad
  \textsuperscript{2}Microsoft \quad
  \textsuperscript{3}KAIST
}

\begin{document}

\maketitle

\makeatletter
\ifacl@anonymize\else
  \begingroup
  \renewcommand{\thefootnote}{\fnsymbol{footnote}}
  \footnotetext[1]{Work done during an internship at Microsoft Research.}
  \footnotetext[3]{Corresponding authors.}
  \endgroup
\fi
\makeatother

\begin{abstract}
Document editing is among the most common knowledge-work tasks, yet existing benchmarks for evaluating autonomous agents on Word document editing provide fully specified attribute-modification instructions. They do not evaluate agents' ability to infer the appropriate attributes from document context, a capability essential in real-world editing where users expect consistency without spelling out every attribute. We introduce \textbf{\ours{}}, a benchmark comprising 280 base document-editing instances, each paired with three instruction conditions: \emph{descriptive} (the instruction explicitly lists every attribute change), \emph{referential} (the instruction points to another object as an exemplar), and \emph{inferential} (the agent must examine document context to infer what attributes to apply). To evaluate contextual attribute inference, each base instance is constructed so that the target formatting is recoverable from an in-document exemplar. The instances are derived from real Word documents collected from Common Crawl and cover 13 object types and 81 attributes. We evaluate six recent coding-agent systems and find that the best-performing evaluated system, Claude Code with Opus 4.6, achieves a 62.9\% task-level pass rate on descriptive variants but only 40.7\% on inferential variants, indicating that current systems complete tasks less reliably when the required formatting changes must be determined from document context.
\end{abstract}

\section{Introduction}

Document editing is one of the most common knowledge-work tasks. In many scenarios, users expect an object to match existing document formatting without specifying every attribute change---for example, after inserting a new shape or pasting content. As agents increasingly operate computer applications autonomously, evaluating their document editing capabilities becomes an important research question.

Recent benchmarks have begun to evaluate agents on document-processing tasks \cite{xie2024osworld,bonatti2024windowsagentarena,wang2024officebench,chen2025odysseybench}. However, their coverage of Word editing tasks remains incomplete. Existing tasks provide fully specified formatting instructions (e.g., ``change the font to Times New Roman''). Such tasks do not evaluate whether an agent can infer the required formatting changes from document context.

\begin{figure*}[t]
\centering
\includegraphics[width=0.85\textwidth]{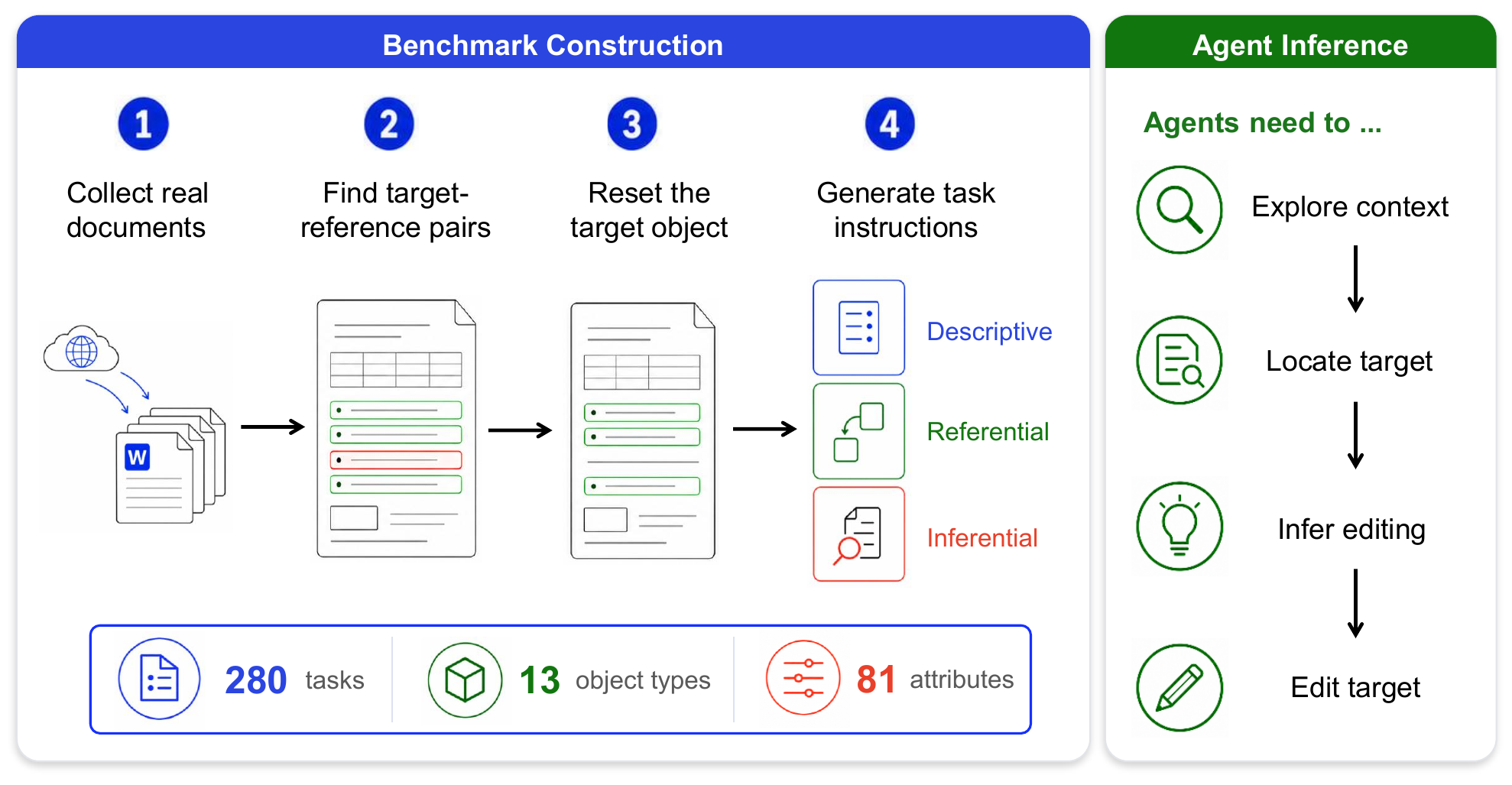}
\caption{Overview of \ours{}. \textbf{Left}: Tasks are constructed from real Word documents by identifying same-type objects with identical formatting, resetting the target's attributes to defaults, and generating three instruction variants. \textbf{Right}: To solve tasks, agents explore context, locate the target, infer the required changes, and execute the edits.}
\label{fig:pipeline}
\end{figure*}

We introduce \textbf{\ours{}}, a benchmark of 280 base editing instances and 840 instruction-conditioned variants designed to evaluate agents on Word document formatting. The benchmark covers 13 object types and 81 distinct attributes. Each base instance contains an input document in which a target object's attributes have been reset to defaults and a held-out reference document representing the desired state. At evaluation time, the agent receives only the input document and an instruction and must restore the target object's attributes; the reference document is used only for scoring. We define three instruction conditions reflecting different user needs and types of available information:

\begin{itemize}
    \item \textbf{Descriptive}: The user specifies every attribute change explicitly (e.g., ``For the picture on page 1, set brightness to 0.85, contrast to 0.15, and set the fill type to solid'').
    \item \textbf{Referential}: The user asks the agent to make one object consistent with another (e.g., ``Modify the picture on page 6 to be consistent with the picture on page 3'') without listing attributes.
    \item \textbf{Inferential}: The user expects the agent to determine what changes are needed from context (e.g., ``Review the document and determine what modification is needed for the header in section 3, then apply the appropriate change'').
\end{itemize}

Together, these conditions evaluate how agents perform formatting tasks under different information conditions. In descriptive tasks, the agent must parse the specifications and execute the attribute changes correctly. In referential tasks, the agent must determine which attributes are relevant for consistency and ensure complete attribute coverage. In inferential tasks, the agent must explore the document context and infer both which existing object serves as the formatting reference and which attributes need to be modified.

For construction, we use real Word documents from Common Crawl \cite{commoncrawl} as reference documents and reset the attributes of a selected object to their defaults to produce the input document. The object's original attributes serve as the ground truth. We require each document to contain at least one other same-type object with an identical formatting signature, ensuring that the original formatting is exactly recoverable from an in-document exemplar. This design limits inferential tasks to same-type exemplar discovery, providing agents with a relatively simple inference setting. Human verification further confirms that the intended formatting is inferable from context and that the reference is unambiguous.

We evaluate six recent coding-agent systems, including Claude Code with Opus 4.6 and Sonnet 4.6, Codex with GPT-5.5 and GPT-5.4, Claude Code with GLM 5.2, and Kimi CLI with Kimi K2.6~\cite{claudecode,claudeopus46,claudesonnet46,openaicodex,gpt55system,gpt54thinking,glm52system,kimicli,kimi26system}, in sandboxed Windows environments. Even on descriptive variants, where all intended attribute changes are explicitly specified, the best evaluated system achieves only a 62.9\% pass rate. All evaluated systems have lower pass rates on inferential variants (best: 40.7\%), indicating that current systems complete tasks less reliably when the required formatting changes must be determined from document context.

Our contributions are as follows:
\begin{itemize}
    \item We propose \ours{}, a benchmark with three paired instruction conditions: explicit specification, reference-based consistency, and context-driven inference.
    \item Our benchmark covers 13 object types and 81 attributes, spanning paragraphs, tables, headers/footers, and various shape types.
    \item We evaluate six recent coding-agent systems, showing that format execution remains difficult for complex objects, and contextual exemplar identification is challenging.
\end{itemize}

\section{Related Work}

\begin{table*}[t]
\centering
{\small
\setlength{\tabcolsep}{3pt}
\begin{tabular}{llcccc}
\toprule
Benchmark & Task Scenario & Word Tasks & Object Types & Attributes & \shortstack{Contextual\\Inference} \\
\midrule
OSWorld & Explicit formatting \& content editing & 47 & 4 & 12 & \xmark \\
OfficeBench & Cross-app workflow & 80 & 1 & 0 & \xmark \\
DocFormEval & Explicit formatting & 1,911 & 1 & 19 & \xmark \\
WorldGUI & Explicit formatting \& content editing & 63 & 7 & 7 & \xmark \\
\textbf{Ours} & \textbf{Context-aware formatting modification} & \textbf{280} & \textbf{13} & \textbf{81} & \cmark \\
\bottomrule
\end{tabular}}
\caption{Comparison of Word-related task coverage across open-source computer-use and document-formatting benchmarks. Contextual Inference denotes tasks requiring agents to infer the required formatting from document context.}
\label{tab:comparison}
\end{table*}

\subsection{Word-Related Evaluation}

Recent benchmarks evaluate autonomous agents in real computer environments, where Word-related tasks are common given the ubiquity of document processing in daily workflows.

OSWorld \cite{xie2024osworld} constructs tasks spanning multiple desktop applications, testing agents' ability to complete diverse computer tasks through GUI interactions. Windows Agent Arena \cite{bonatti2024windowsagentarena} adapts the OSWorld framework to a scalable Windows environment. WorldGUI \cite{chen2024worldgui} benchmarks desktop GUI automation under varied initial states across ten applications. OfficeBench \cite{wang2024officebench} focuses on multi-application workflow orchestration across office tools. OdysseyBench \cite{chen2025odysseybench} evaluates long-horizon workflows that require agents to recover information from interaction histories and act across office applications. DocFormEval \cite{rao2024docformeval} benchmarks code generation for Word formatting from natural-language instructions.

More recently, DocFormBench \cite{rao2026docformbench} proposes 500 Chinese and English instances for content-aware Word formatting, where agents must locate target elements via natural-language descriptions and apply explicitly specified format changes. OfficeEval \cite{lv2026officeeval} constructs 200 tasks drawn from standardized Office proficiency exams with 7,118 machine-scorable rubrics, covering fine-grained formatting attributes in Word and other applications.

Despite their contributions, these benchmarks remain limited in evaluating formatting inference from document context under different information conditions. \ours{} addresses this limitation by constructing formatting inference tasks with descriptive, referential, and inferential instruction variants.

\subsection{Document Editing}

A related line of work studies automated document editing driven by natural-language instructions. DocEdit-v2 \cite{suri2024doceditv2} grounds edit requests to regions in document images and generates editing commands via a multimodal LLM. Agent-DocEdit \cite{wu2024agentdocedit} localizes target elements in content-rich documents and generates edit programs through pre-defined APIs. LEDGER \cite{wang2026ledger} uses dependency-aware graphs and graph-guided retrieval for semantically consistent edits across long-document paragraphs.

These works address semantic or layout editing, target grounding, and cross-section consistency under high-level instructions. They do not evaluate exact recovery of native Word formatting attributes from in-document exemplars, which is the focus of our benchmark.

\subsection{Benchmarks for Other Office Applications}

Benchmarks have also been developed for other Office applications. In the presentation domain, SlidesGen-Bench \cite{yang2026slidesgenbench} evaluates automated slide generation across content, aesthetics, and editability, while PresentBench \cite{chen2026presentbench} provides instance-specific, fine-grained rubric evaluation. In the spreadsheet domain, FINCH (also introduced as FinWorkBench) \cite{dong2025finworkbench} targets enterprise finance and accounting workflows. SpreadsheetBench \cite{ma2024spreadsheetbench} collects real user questions from online forums and tests spreadsheet manipulation.

Spreadsheet benchmarks often require agents to inspect file structure before acting. However, existing Word benchmarks do not specifically evaluate tasks requiring target formatting to be inferred from within the document. \ours{} evaluates this capability through context-driven formatting tasks.

\section{Benchmark Construction}

\subsection{Task Formulation}

In \ours{}, each base instance consists of an input document, a held-out reference document, and three editing-instruction variants. The agent receives only the input document and one instruction; the reference document is not exposed and is used solely for scoring. The instruction requires modifying a target object in the input document to match its state in the reference document.

Tasks are divided into three instruction specificity conditions based on the amount of information provided in the instruction:

\begin{itemize}
    \item \textbf{Descriptive}: The instruction describes the specific attribute changes in detail. The agent only needs to understand and execute each change.
    \item \textbf{Referential}: The instruction specifies another object of the same type in the document as a reference, without describing the specific changes. The agent must read the reference object's attributes and apply them to the target.
    \item \textbf{Inferential}: The instruction only identifies the target object, providing no reference or specific description. The agent must browse the document context and infer what changes should be made.
\end{itemize}

\begin{figure*}[t]
\centering
\resizebox{0.9\textwidth}{!}{%
\begin{minipage}{\textwidth}
\centering
\begin{tikzpicture}
\definecolor{instrubg}{HTML}{F3F6FA}
\definecolor{instrubar}{HTML}{285A9F}
\definecolor{instuborder}{HTML}{D8DEE8}
\node[inner sep=5pt, inner xsep=8pt, fill=instrubg, text width=\textwidth-18pt, anchor=north west, rounded corners=2pt] (inst) {%
\small\sffamily\color{instrubar}\textbf{Instruction}\\[1pt]
\color{black}\textrm{\footnotesize Review the document and determine the appropriate formatting adjustments needed for the AutoShape named ``AutoShape 3'' on page 1. Apply only formatting changes to this object, keeping all document content and text exactly the same.}};
\fill[instrubar, rounded corners=1.5pt] (inst.north west) rectangle ([xshift=3pt]inst.south west);
\end{tikzpicture}

\vspace{3pt}

\begin{minipage}[t]{0.485\textwidth}
    \centering
    \docframe{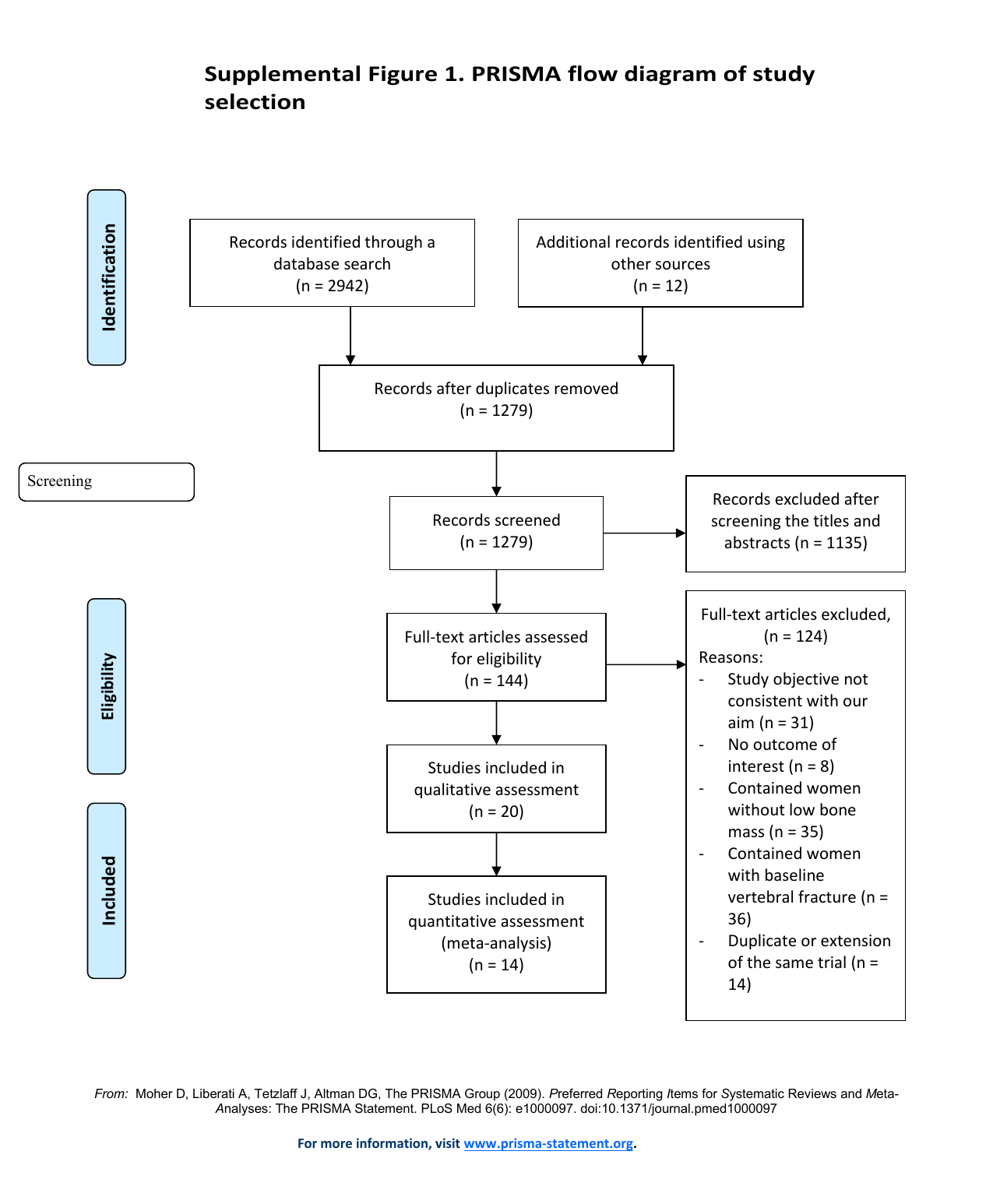}{Input document}
\end{minipage}
\hfill
\begin{minipage}[t]{0.485\textwidth}
    \centering
    \docframe{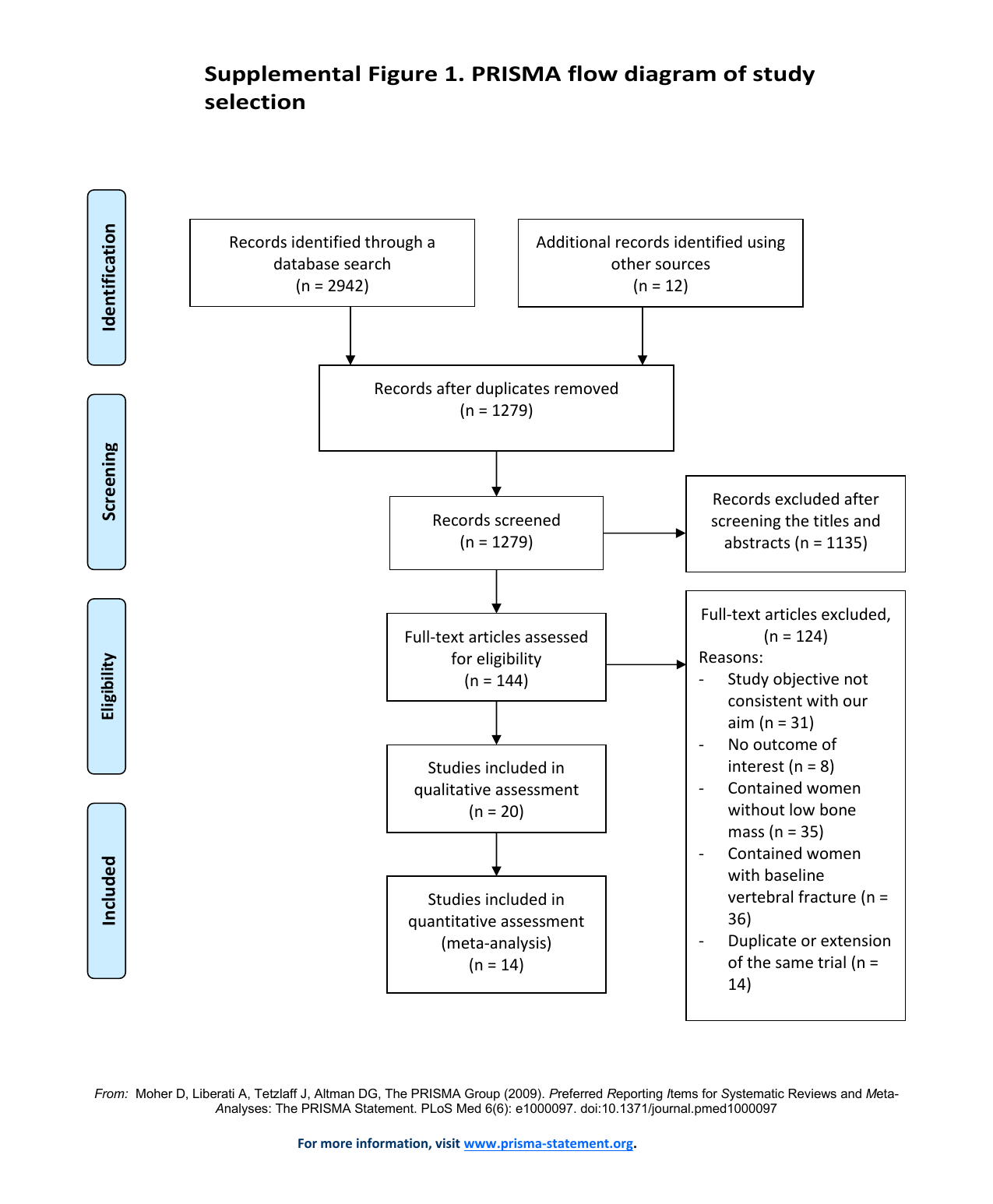}{Reference document}
\end{minipage}

\vspace{3pt}

\begin{tikzpicture}
\definecolor{instrubg}{HTML}{F3F6FA}
\definecolor{instrubar}{HTML}{285A9F}
\definecolor{chipbg}{HTML}{DCE5F0}
\node[inner sep=5pt, inner xsep=8pt, fill=instrubg, text width=\textwidth-18pt, anchor=north west, rounded corners=2pt] (attrs) {%
\small\sffamily\color{instrubar}\textbf{Evaluated Attributes}\\[-2pt]
\footnotesize
\tikz[baseline=-0.5ex]\node[fill=chipbg, rounded corners=2pt, inner sep=3pt, inner xsep=6pt]{\color{black!80}\scriptsize Rotation};\enspace%
\tikz[baseline=-0.5ex]\node[fill=chipbg, rounded corners=2pt, inner sep=3pt, inner xsep=6pt]{\color{black!80}\scriptsize Text Orientation};\enspace%
\tikz[baseline=-0.5ex]\node[fill=chipbg, rounded corners=2pt, inner sep=3pt, inner xsep=6pt]{\color{black!80}\scriptsize Fill Color};\enspace%
\tikz[baseline=-0.5ex]\node[fill=chipbg, rounded corners=2pt, inner sep=3pt, inner xsep=6pt]{\color{black!80}\scriptsize Style};\enspace%
\tikz[baseline=-0.5ex]\node[fill=chipbg, rounded corners=2pt, inner sep=3pt, inner xsep=6pt]{\color{black!80}\scriptsize Alignment};\enspace%
\tikz[baseline=-0.5ex]\node[fill=chipbg, rounded corners=2pt, inner sep=3pt, inner xsep=6pt]{\color{black!80}\scriptsize Keep With Next};\enspace%
\tikz[baseline=-0.5ex]\node[fill=chipbg, rounded corners=2pt, inner sep=3pt, inner xsep=6pt]{\color{black!80}\scriptsize Bold};\enspace%
\tikz[baseline=-0.5ex]\node[fill=chipbg, rounded corners=2pt, inner sep=3pt, inner xsep=6pt]{\color{black!80}\scriptsize Font};\enspace%
\tikz[baseline=-0.5ex]\node[fill=chipbg, rounded corners=2pt, inner sep=3pt, inner xsep=6pt]{\color{black!80}\scriptsize Font Size};%
};
\fill[instrubar, rounded corners=1.5pt] (attrs.north west) rectangle ([xshift=3pt]attrs.south west);
\end{tikzpicture}
\end{minipage}}
\caption{A complete test case from \ours{}. The inferential instruction specifies only the target object; the agent must infer what attribute modifications are needed from document context. \textbf{Left}: the input document whose target object has been reset to defaults. \textbf{Right}: the held-out reference document with original formatting. The evaluator compares the 9 listed attributes between the agent's output and the reference document.}
\label{fig:example-task}
\end{figure*}

\subsection{Construction Method}

We collect real Word documents from Common Crawl \cite{commoncrawl} as reference documents. These documents originate from heterogeneous web sources and contain varied typographic structures. We reset the attributes of a selected object in a copy of a real document to construct the input document, while the object's original attributes define the evaluation ground truth. We select objects for which the context contains at least one exactly matching same-type exemplar. If an object's formatting appears only once in the document, its original attributes are not recoverable from context after resetting, so it is ineligible.

To filter documents and objects that can produce valid tasks, we first parse the XML of Word documents to identify their object types, retain documents containing at least two objects of a given type, and sample approximately 1,000 documents per category as candidates. For each candidate document, we extract the original attributes of all objects via Word COM. We then reset each candidate object independently in a separate document copy and compute the difference between its reset and original states.

On this basis, we pair objects of the same type within the same document that share identical formatting but differ in content, forming (target, reference) pairs. This construction scopes our inferential tasks to same-type exemplar discovery and exact attribute restoration. Exact matching ensures that the original formatting is recoverable from at least one in-document exemplar and enables deterministic evaluation through exact attribute comparison. We further use human verification to confirm that the intended formatting is inferable from context and that the reference answer is unambiguous (Section~\ref{sec:human_verification}). While real-world scenarios may also involve inferring from multiple objects or coordinating changes across multiple targets, these are beyond the current scope. To cover as many attributes as possible, we employ a greedy set-cover algorithm to sample from all candidate pairs, maximizing the variety of modified attributes covered: iteratively selecting the object whose attribute differences cover the most uncovered attributes, breaking ties by preferring objects that cover more low-frequency attributes, and limiting each document to contributing at most one object per type.

We use GPT-5.5 to generate three instruction variants for each base instance (see Appendix~\ref{sec:data_gen_prompts} for full prompts). For descriptive instructions, we first compute the target object's structured attribute diff, converting all raw COM values to human-readable representations, e.g., enum constants to names (fill type \texttt{3} becomes \texttt{"Gradient"}). This structured diff is fed to the LLM to rewrite all changes as natural-language instructions, followed by a second refinement pass for conciseness. For referential instructions, the LLM generates instructions requiring format consistency between target and exemplar objects based solely on their positional information (type, page number, content), without specifying attributes. Inferential instructions provide only the target object's positional information, requiring the agent to infer the modification plan from document context.

\subsection{PII Removal}

To protect privacy, we perform PII detection at both the image and text levels. Documents are rendered as page screenshots and an LLM detects face photographs; documents containing faces are entirely removed. Document text is scanned by an LLM for 8 categories of PII (see Appendix~\ref{sec:pii_filtering}), and detected PII text is replaced with placeholders. After replacement, we verify that the target object's attributes are unaffected; otherwise the entry is removed.

\subsection{Human Verification}
\label{sec:human_verification}

We recruit 9 annotators divided into 3 groups of 3 to verify all samples (see Appendix~\ref{sec:human_annotation} for the detailed annotation protocol). Each group annotates all samples, where each sample is assigned to exactly one instruction type within that group. The assignment is rotated across groups so that every sample's three instruction types are each verified by one group. To verify inferability and answer uniqueness, annotators first complete the edit using only the input document and instruction, without access to the held-out reference. They then inspect the reference and rate instruction clarity, reference correctness, and agreement between their independently produced output and the reference. This blind-first procedure provides an independent check that the intended formatting is inferable from context and that the reference answer is unambiguous. Annotators also check for PII. A sample is removed if all 3 annotators in at least one group reach consensus that it is problematic on any dimension. Cases with non-unanimous judgments are manually reviewed based on the annotators' feedback, and the instruction or reference is revised when necessary. After this process, 280 of the original 312 candidates are retained (89.7\%).

\subsection{Statistics}

The benchmark contains 280 base editing instances, each paired with descriptive, referential, and inferential instructions, yielding 840 instruction-conditioned variants across 272 distinct documents. It covers 81 attributes across 13 object types: plain paragraph, listing, hyperlink, table of contents, table, header, footer, autoshape, textbox, picture, chart, SmartArt, and WordArt (see Appendix~\ref{sec:benchmark_stats} for details). These attributes include character and paragraph formatting, hyperlinks, table borders and cell formatting, shape fill, line, shadow, 3D, glow, reflection, and other object-specific properties.

The statistics of the 280 inferential variants highlight why this setting requires more than local attribute copying. In 42.9\% of inferential variants, the selected matching exemplar is on a different page from the target, so the relevant evidence can extend beyond the target page. Documents also contain many same-type distractors: each instance contains a median of 11.5 same-type objects, with a median non-matching ratio of 80.9\%. Thus, inferential variants require agents to decide which contextual objects are relevant before determining the attributes to edit.

\section{Results and Analysis}

\subsection{Experimental Setup}

We evaluate the benchmark on six coding-agent systems: Claude Code~\cite{claudecode} with Opus 4.6~\cite{claudeopus46}, Claude Code with Sonnet 4.6~\cite{claudesonnet46}, Codex~\cite{openaicodex} with GPT-5.5~\cite{gpt55system}, Codex with GPT-5.4~\cite{gpt54thinking}, Claude Code with GLM 5.2~\cite{glm52system}, and Kimi CLI~\cite{kimicli} with Kimi K2.6~\cite{kimi26system} (see Appendix~\ref{sec:agent_config} for exact versions and environment configuration). Each system runs in an isolated Windows sandbox. A fresh sandbox is provisioned for each variant run; no execution history, cache, or workspace state is shared between runs, including among variants of the same base instance. Each system autonomously chooses its editing approach and executes modifications; the output document is scored against the held-out reference through Word COM.

We use two metrics: \textbf{Task-Level Pass Rate}, where a task passes only when all evaluated changed attributes of the target object are correct; and \textbf{Attribute-Level Accuracy}, which measures the fraction of target attributes requiring modification that are correctly updated in each task, macro-averaged across tasks. The evaluator compares the target attributes that differ between the input and reference documents. Both the reference and prediction documents are processed by the same COM-based extraction pipeline, which normalizes attribute values into canonical string representations with standardized units, maps COM enum constants to human-readable names, and treats sentinel values such as \texttt{wdUndefined} as absent. Because extraction reads the effective property values resolved by Word's rendering engine, attributes set via style inheritance and direct formatting are compared equivalently. Prediction files that cannot be opened are scored as zero.

Beyond target correctness, we separately evaluate whether model outputs preserve non-target document content. Target correctness and non-target preservation are scored independently. The preservation evaluator compares supported non-target objects and their contents and formatting between the reference and model-generated documents. An output passes the preservation check only when no differences are detected in any evaluated non-target object.

\begin{table*}[t]
\centering
{\small
\setlength{\tabcolsep}{3pt}
\begin{tabular}{lcccccccc}
\toprule
 & \multicolumn{2}{c}{Descriptive} & \multicolumn{2}{c}{Referential} & \multicolumn{2}{c}{Inferential} & \multicolumn{2}{c}{Average} \\
\cmidrule(lr){2-3} \cmidrule(lr){4-5} \cmidrule(lr){6-7} \cmidrule(lr){8-9}
Agent & Pass Rate & Attr Acc & Pass Rate & Attr Acc & Pass Rate & Attr Acc & Pass Rate & Attr Acc \\
\midrule
Human Baseline & 50.0\% & 85.1\% & 43.3\% & 75.3\% & 34.0\% & 70.1\% & 42.5\% & 76.9\% \\
\midrule
Claude Code (Opus 4.6) & 62.9\% & 82.3\% & 77.1\% & 87.5\% & 40.7\% & 59.3\% & 60.2\% & 76.4\% \\
Claude Code (Sonnet 4.6) & 45.4\% & 78.9\% & 56.8\% & 71.1\% & 29.3\% & 48.2\% & 43.8\% & 66.1\% \\
Codex (GPT-5.5) & 55.4\% & 70.0\% & 43.6\% & 58.7\% & 25.7\% & 43.7\% & 41.5\% & 57.5\% \\
Claude Code (GLM 5.2) & 31.4\% & 51.8\% & 56.4\% & 67.8\% & 20.0\% & 35.9\% & 36.0\% & 51.8\% \\
Codex (GPT-5.4) & 41.4\% & 60.4\% & 32.1\% & 45.0\% & 23.2\% & 38.0\% & 32.3\% & 47.8\% \\
Kimi CLI (Kimi K2.6) & 37.5\% & 68.2\% & 41.8\% & 63.4\% & 15.4\% & 28.8\% & 31.5\% & 53.5\% \\
\bottomrule
\end{tabular}}
\caption{Overall results across three instruction conditions. Pass Rate is the proportion of tasks in which all evaluated changed attributes are correct. Attr Acc is the fraction of target attributes requiring modification that are correctly updated in each task, macro-averaged across tasks. Both metrics evaluate only the target object; non-target preservation is reported separately.}
\label{tab:overall}
\end{table*}

\begin{figure*}[t]
\centering
\includegraphics[width=\textwidth]{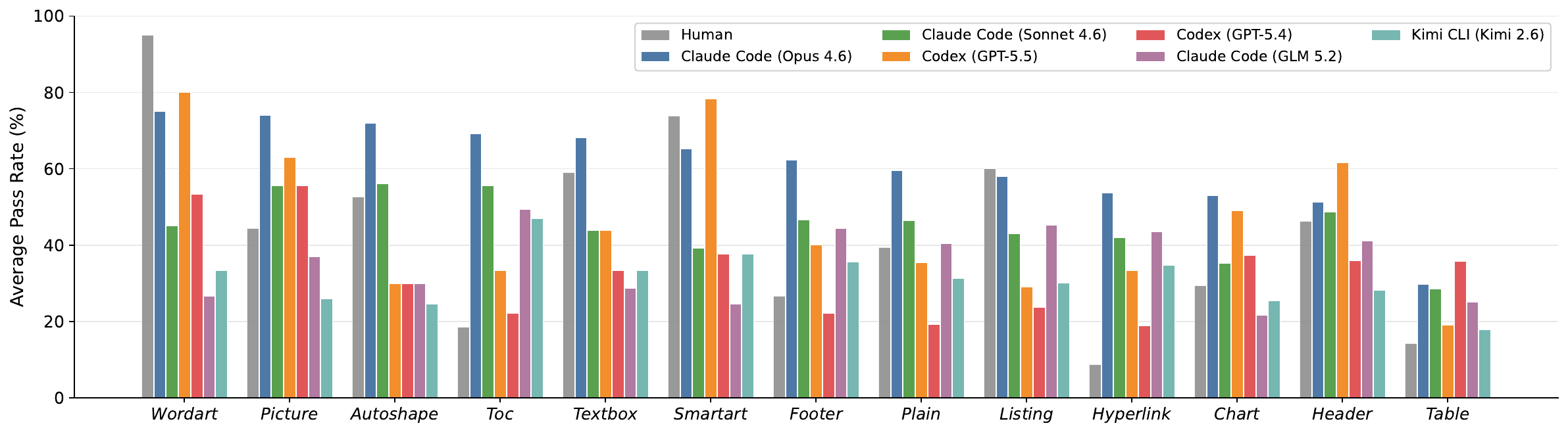}
\caption{Average task-level pass rates (\%) by object type across all three instruction conditions.}
\label{fig:perobject}
\end{figure*}

\subsection{Evaluation Results}

Table~\ref{tab:overall} presents the overall results. All evaluated agent systems obtain lower pass rates on inferential variants than on descriptive and referential variants. Even the best-performing system, Claude Code with Opus 4.6, achieves only a 40.7\% pass rate on inferential variants.

Notably, several agents achieve higher pass rates on referential tasks than on descriptive tasks, though this pattern does not hold uniformly across all agents. One possible explanation is that programmatically copying attributes from a reference object can be simpler than parsing detailed natural-language attribute specifications, which requires correctly interpreting human-readable attribute names and values and mapping them to exact API parameters. The inconsistency across agents may reflect differences in their natural-language instruction following and code generation capabilities.

Figure~\ref{fig:perobject} shows average task-level pass rates by object type. Tables have the lowest pass rates across all agents, as they require modifying attributes across multiple cells, where each cell contains border properties, shading, vertical alignment, and text formatting, creating a deep attribute hierarchy that agents struggle to handle completely. Per-attribute accuracy results are provided in Table~\ref{tab:attr_list} of Appendix~\ref{sec:benchmark_stats}.

We also include a human baseline (Table~\ref{tab:overall}), where 9 annotators complete tasks under the same conditions as agents: given an instruction and an input document, they manually edit the document in Microsoft Word without access to the reference document, and the output is scored using the identical automated evaluation pipeline (see Appendix~\ref{sec:human_annotation} for annotation protocol details). Analyzing human errors, we find that annotators generally execute individual attribute modifications correctly but tend to overlook some of the required attributes. As a result, their attribute-level accuracy is comparable to or slightly better than the best agent (76.9\% vs.\ 76.4\%), yet their task-level pass rate is lower (42.5\% vs.\ 60.2\%), since pass rate requires all attributes to be correct and a single omission fails the entire task. Appendix~\ref{sec:common_target_errors} compares common attribute errors in human and agent outputs.

Preservation results show that collateral changes can occur even when editing is directed at a specific target. For example, Claude Code with Opus 4.6 achieves a preservation pass rate of 89.6\% under the inferential condition. Common failure types include changes to textual content, formatting changes propagating to unrelated objects, and structural changes to non-target objects. Detailed results by model and instruction condition are provided in Appendix~\ref{sec:preservation_results}.

\subsection{Sources of Difficulty in Inferential Editing}

To probe different sources of difficulty, we conduct two diagnostic experiments using Codex with GPT-5.5. The exemplar-selection diagnostic requires only an object selection and is evaluated with localization accuracy, whereas the property-path diagnostic performs document editing and is evaluated by exact target-attribute matching.

First, we isolate exemplar localization, one component of contextual inference. GPT-5.5 receives the input document and inferential instruction but performs no document modification and is not given a candidate list. It selects a same-type object as the formatting exemplar. If multiple same-type objects have the required target formatting, selecting any of them is counted as correct. GPT-5.5 achieves an exemplar-selection accuracy of 81.2\%. This diagnostic measures only exemplar localization, not property-path identification or document editing. This result indicates that exemplar inference itself remains a source of error even when editing execution is excluded.

Second, starting from the referential setting, where the exemplar is explicitly provided, we additionally provide the property paths to be modified without revealing their target values. The agent must infer the values from the exemplar and produce the edited document. Providing the property paths increases the task-level pass rate from 43.6\% to 45.4\%. Thus, under this setting, explicit property paths yield only a small improvement in target-object modification. Separately, in the descriptive condition, both the properties and their target values are explicitly specified, yet the same agent achieves a pass rate of only 55.4\%. This shows that failures remain even when contextual exemplar and property discovery are not required. Taken together, these results indicate that inferential-task failures reflect both errors in exemplar localization and failures to correctly apply the required formatting changes.

\subsection{Failure Analysis}

Qualitative inspection of failed agent trajectories revealed the following illustrative failure modes:

\paragraph{Improper editing methods causing file corruption or ineffective modifications.} A Word document is essentially a ZIP archive containing XML files. In inspected failures, some agents unzip the document, modify the internal XML, and repackage it rather than editing through Word's programming interface. This approach can produce a malformed package even when the local XML edit appears correct. For example, an XML library may rename namespace prefixes during re-serialization while other locations still reference the original names, causing broken references. Word may then reject the file or silently discard modifications.

\paragraph{Incorrect understanding of API parameter values or semantics.} When executing attribute modifications through programming interfaces, agents sometimes misunderstand parameter values or meanings. For example, when setting chart type to ``Scatter,'' an agent used enum value 74, which actually corresponds to ``Scatter with Straight Lines and Markers,'' while the correct value for a plain scatter plot is $-4169$. Similarly, when setting both paragraph left indent and hanging indent simultaneously, agents confuse the meaning of the \texttt{LeftIndent} property, which actually represents the sum of left indent and hanging indent, not the left indent alone.

\paragraph{Incomplete attribute coverage.} In referential and inferential tasks, agents must identify which attributes to modify on their own. Selected trajectories show cases where an agent locates the appropriate exemplar but omits some required attributes. The observed examples include complex nested objects for which the agent modifies outer attributes but neglects per-run formatting within paragraphs, per-cell styles in tables, or internal text formatting within shapes. Descriptive instructions specify these attributes, removing the discovery requirement, though execution may still fail.

\subsection{Limited Visual Understanding of Document Formatting}

In referential and inferential tasks, agents must autonomously determine which attributes of the target object need to be modified. A direct approach is to visually inspect the document context, comparing the target object with other same-type objects to infer the required attribute changes. We designed an experiment to test whether GPT-5.5 can identify formatting attribute changes through visual inspection.

For each of the 280 input--reference pairs, we render screenshots of up to three differing pages per document and ask GPT-5.5 to select the changed attributes from an 81-attribute checklist. We compare its selections with programmatically computed ground-truth attribute diffs. This setting is easier than the actual task, where agents have access to neither the reference document nor an attribute checklist.

GPT-5.5 achieves 0.740 precision, 0.351 recall, and 0.476 F1: its predictions are mostly correct but recover only about one-third of the changes, often missing visually subtle properties such as line spacing, paragraph spacing, and font. Figure~\ref{fig:visual_example} in Appendix~\ref{sec:visual_example_appendix} shows an example, consistent with prior findings on fine-grained visual perception \cite{rahmanzadehgervi2024vision,liu2024focusanywhere}. Because coding agents may also inspect document properties through COM or XML, this result does not explain all agent failures, but cautions against relying on screenshot-only attribute detection or vision-only LLM judges without validation.

\section{Conclusion}

We present \ours{}, a benchmark of 280 base editing instances and 840 instruction-conditioned variants that evaluates coding-agent systems on Word document formatting under descriptive, referential, and inferential instructions. Across six evaluated systems, pass rates are lowest on inferential variants. Qualitative trajectory inspection identifies illustrative bottlenecks including improper editing methods, misunderstanding of API parameter semantics, and incomplete attribute coverage, while a controlled GPT-5.5 experiment shows that screenshot-only inspection misses many fine-grained formatting differences. These findings show that context-aware attribute inference remains a core challenge for current coding-agent systems.

\section*{Limitations}

Our benchmark currently targets only formatting attribute modifications in Word documents, without covering other Office applications such as PowerPoint and Excel. Our three-level instruction design and attribute-comparison-based evaluation framework are in principle transferable to other applications, and extending to a broader range of Office application scenarios is a direction for future work. Additionally, for evaluation convenience, our task design is limited to modifying a single target object per task. In real-world document editing, user requests often involve coordinated modifications across multiple objects, such as uniformly adjusting the formatting of all headings in a document. However, an agent's ability to modify formatting on a single object is the foundation for completing multi-object coordinated edits, and performance on single-object tasks can reflect its core capability in more complex scenarios.

\section*{Ethical Considerations}

We use publicly accessible Word documents collected through Common Crawl. The original documents may contain personally identifiable information (PII). To reduce privacy risks, we replace detected PII with placeholders, exclude documents containing detected face photographs, and conduct additional privacy checks during human verification.

\bibliography{wordbench}

\appendix

\section{Agent Evaluation Configuration}
\label{sec:agent_config}

\subsection{Agent Systems Under Test}

Table~\ref{tab:agent_systems} lists the six agent systems evaluated in our experiments.

\begin{table*}[t]
\centering
\small
\caption{Agent systems and their configurations.}
\label{tab:agent_systems}
\begin{tabular}{@{}lll@{}}
\toprule
\textbf{Agent System} & \textbf{Underlying Model} & \textbf{CLI Tool \& Version} \\
\midrule
Claude Code & Claude Opus 4.6 & \texttt{claude-code@2.1.87} \\
Claude Code & Claude Sonnet 4.6 & \texttt{claude-code@2.1.87} \\
Codex & GPT-5.5 & \texttt{codex@0.138.0} \\
Codex & GPT-5.4 & \texttt{codex@0.138.0} \\
Claude Code & GLM 5.2 & \texttt{claude-code@2.1.87} \\
Kimi CLI & Kimi K2.6 & \texttt{kimi-cli@1.47.0} \\
\bottomrule
\end{tabular}
\end{table*}

\subsection{Execution Environment}

\begin{table}[t]
\centering
\small
\caption{Execution environment specification.}
\label{tab:exec_env}
\begin{tabular}{@{}ll@{}}
\toprule
\textbf{Component} & \textbf{Specification} \\
\midrule
OS & Windows 11 Enterprise 25H2 \\
VM Resources & 8\,GB RAM, 2 cores, 32\,GB disk \\
Per-task Timeout & 600 seconds \\
Runs per Variant & 1; provider-default decoding settings \\
Office & Microsoft Office LTSC 2024 \\
Language / Locale & English (default Windows 11 install) \\
\bottomrule
\end{tabular}
\end{table}

Each task run uses a freshly provisioned sandbox. No execution history, cache, or workspace state is shared between runs. Agents have access to PowerShell, Python, and the Word COM automation interface. No network access is provided beyond the agent's own API endpoint.

\subsection{Task Prompt}

All agents receive the identical task prompt shown below. No additional system prompts, skill libraries, or built-in Office knowledge are injected beyond what each CLI tool provides by default.

\begin{quote}
\small\ttfamily
You are a Word document formatting expert.\\[4pt]
You need to modify a Word document (.docx) according to the given instruction.\\[4pt]
\#\#\# Instruction\\
\{instruction\}\\[4pt]
\#\#\# Document Path\\
\{doc\_path\}\\[4pt]
\#\#\# Output Path\\
\{output\_path\}\\[4pt]
Please follow these steps:\\
1. First, make a copy of the document at the document path to the output path, so you work on the copy.\\
2. Complete the task described in the instruction on the document at the output path.
\end{quote}

Input documents are placed in the agent's sandbox directory with a fixed filename (\texttt{input.docx}). The output path is explicitly specified in the prompt (\texttt{output.docx} in the same directory). Agents do not need to discover or search for files.

\section{Benchmark Details}
\label{sec:benchmark_stats}

Table~\ref{tab:task_stats} shows the number of tasks per object type in the final benchmark after filtering. Table~\ref{tab:attr_list} lists all 81 evaluated attributes with the number of tasks testing each attribute ($N$) and the available per-attribute accuracy averaged across all six agents.

The benchmark documents are selected from Common Crawl content. Copyright in third-party documents and their contents remains with the respective rights holders; the benchmark release does not grant additional rights to these materials. Project-authored task instructions, annotations, and metadata are released under the MIT License.

\begin{table}[t]
\centering
\small
\caption{Number of tasks per object type.}
\label{tab:task_stats}
\begin{tabular}{@{}lr@{}}
\toprule
\textbf{Object Type} & \textbf{Tasks} \\
\midrule
Plain paragraph & 33 \\
Listing & 31 \\
Table & 28 \\
Table of contents & 27 \\
Hyperlink & 23 \\
SmartArt & 23 \\
Textbox & 22 \\
WordArt & 20 \\
Autoshape & 19 \\
Chart & 17 \\
Footer & 15 \\
Header & 13 \\
Picture & 9 \\
\midrule
\textbf{Total} & \textbf{280} \\
\bottomrule
\end{tabular}
\end{table}

\begin{table*}[p]
\centering
\scriptsize
\caption{All 81 evaluated formatting attributes with per-attribute accuracy averaged across all six agents. $N$ = number of tasks (out of 280) that evaluate each attribute.}
\label{tab:attr_list}
\begin{tabular}{@{}lrr@{\hspace{1.5em}}lrr@{}}
\toprule
\textbf{Attribute} & \textbf{$N$} & \textbf{Acc\%} & \textbf{Attribute} & \textbf{$N$} & \textbf{Acc\%} \\
\midrule
font\_formatting.bold & 108 & 65.5 & contrast & 2 & 55.6 \\
font\_formatting.color & 96 & 53.8 & data\_table\_show\_legend\_key & 7 & 41.3 \\
font\_formatting.font & 148 & 59.6 & fill\_back\_color\_rgb & 10 & 63.9 \\
font\_formatting.font\_size & 155 & 64.1 & fill\_fore\_color\_rgb & 40 & 61.9 \\
font\_formatting.italic & 49 & 64.4 & fill\_gradient\_angle & 7 & 71.4 \\
font\_formatting.strike & 1 & 50.0 & fill\_transparency & 15 & 58.5 \\
font\_formatting.subscript & 6 & 52.8 & fill\_type & 18 & 61.7 \\
font\_formatting.superscript & 14 & 56.0 & glow\_color\_rgb & 1 & 61.1 \\
font\_formatting.underline & 63 & 59.0 & glow\_radius & 1 & 61.1 \\
paragraph\_formatting.alignment & 91 & 64.8 & has\_data\_table & 7 & 41.3 \\
paragraph\_formatting.indent.hanging & 82 & 61.9 & has\_legend & 10 & 53.3 \\
paragraph\_formatting.indent.left & 86 & 46.0 & has\_title & 11 & 48.5 \\
paragraph\_formatting.indent.right & 37 & 62.6 & layout\_name & 23 & 48.3 \\
paragraph\_formatting.keep\_lines\_together & 47 & 68.9 & line\_dash\_style & 4 & 50.0 \\
paragraph\_formatting.keep\_with\_next & 52 & 69.1 & line\_fore\_color\_rgb & 32 & 59.4 \\
paragraph\_formatting.line\_spacing & 110 & 63.0 & line\_visible & 20 & 63.1 \\
paragraph\_formatting.line\_spacing\_rule & 107 & 59.9 & line\_weight & 35 & 62.9 \\
paragraph\_formatting.list.level & 61 & 59.3 & quick\_style\_name & 19 & 48.5 \\
paragraph\_formatting.list.type & 62 & 57.3 & reflection\_blur & 3 & 59.3 \\
paragraph\_formatting.page\_break\_before & 10 & 71.7 & reflection\_offset & 3 & 59.3 \\
paragraph\_formatting.spacing\_after & 127 & 64.5 & reflection\_size & 3 & 51.9 \\
paragraph\_formatting.spacing\_before & 89 & 65.2 & reflection\_transparency & 3 & 59.3 \\
paragraph\_formatting.style & 158 & 63.5 & rotation & 6 & 51.9 \\
alignment & 8 & 59.7 & series\_info.data\_label\_position & 6 & 50.9 \\
axes.has\_major\_gridlines & 13 & 48.7 & series\_info.has\_data\_labels & 6 & 53.7 \\
axes.has\_minor\_gridlines & 2 & 44.4 & series\_info.has\_error\_bars & 2 & 33.3 \\
brightness & 2 & 55.6 & series\_info.line\_color\_rgb & 15 & 48.5 \\
cell\_formatting.border.color & 27 & 47.7 & shadow\_angle & 29 & 60.2 \\
cell\_formatting.border.exists & 26 & 51.9 & shadow\_distance & 30 & 61.3 \\
cell\_formatting.border.line\_style & 13 & 52.1 & shadow\_visible & 19 & 62.3 \\
cell\_formatting.border.line\_width & 22 & 50.8 & soft\_edge\_radius & 1 & 44.4 \\
cell\_formatting.shading & 22 & 52.8 & style & 33 & 53.4 \\
hyperlink & 48 & 57.3 & text\_frame\_orientation & 2 & 44.4 \\
cell\_formatting.vertical\_alignment & 17 & 51.3 & threed\_bevel\_top\_depth & 1 & 66.7 \\
cell\_margin & 19 & 42.7 & threed\_bevel\_top\_type & 1 & 66.7 \\
cell\_spacing & 4 & 30.6 & threed\_depth & 5 & 54.4 \\
chart\_type & 15 & 46.3 & threed\_rotation\_x & 1 & 55.6 \\
color\_name & 19 & 48.8 & threed\_visible & 4 & 50.0 \\
color\_type & 3 & 59.3 & wordart\_kerned\_pairs & 3 & 55.6 \\
column\_widths.width & 11 & 28.8 & wordart\_tracking & 3 & 63.0 \\
column\_widths.width\_rule & 9 & 53.1 & & & \\
\bottomrule
\end{tabular}
\end{table*}

\section{Data Generation Prompts}
\label{sec:data_gen_prompts}

All prompts below are sent to GPT-5.2 as user messages with no system prompt.

\subsection{Instruction Generation (Two-pass)}

\paragraph{Pass 1: Structured changes to natural language.}

\begin{quote}
\small\ttfamily
Rewrite the given document modification changes into a clear, natural language instruction:\\
Requirements:\\
1. Describe every change listed. Do not skip any, even when a value is set to 0 or reset to default.\\
2. When multiple components (e.g., different runs in a paragraph) change differently, describe each one individually.\\
3. When multiple parts share the exact same change, merge them into one statement instead of repeating.\\
4. Only return the instruction. Use plain text without markdown formatting.\\
5. Always write the instruction in English, but keep document content in its original language.
\end{quote}

\paragraph{Pass 2: Refinement.}

\begin{quote}
\small\ttfamily
Refine the following document modification instruction to be more natural and concise:\\
Rules:\\
1. Do not repeat the same location in every sentence. State each location once, then describe all its changes together.\\
2. Do not drop or add any changes---the content must remain exactly the same.\\
3. Only return the rewritten instruction. Use plain text without markdown formatting.\\
4. Always write the instruction in English, but keep document content in its original language.\\
Instruction to refine:
\end{quote}

\subsection{Referential Instruction Generation}

\begin{quote}
\small\ttfamily
Rewrite the given reference-based document modification task into a clear, natural language instruction.\\
Requirements:\\
1. Describe which object needs to be modified and which object is the reference.\\
2. The instruction should ask to make the target object consistent with the reference object.\\
3. Do NOT specify or guess which attributes to change. Just describe the target and reference objects.\\
4. Only return the final instruction text. No markdown, bullets, or headers.\\
5. Always write the instruction in English, but keep document content in its original language.
\end{quote}

\subsection{Inferential Instruction Generation}

\begin{quote}
\small\ttfamily
Rewrite the given document modification task into a clear, natural language instruction.\\
Requirements:\\
1. Describe which object needs to be modified.\\
2. The instruction should ask to review the document and infer the appropriate modification for the target object.\\
3. Do NOT specify or guess which attributes to change.\\
4. Only return the final instruction text. No markdown, bullets, or headers.\\
5. Always write the instruction in English, but keep document content in its original language.
\end{quote}

\section{PII Filtering}
\label{sec:pii_filtering}

PII detection is performed at both image and text levels using GPT-5.2.

\subsection{Text PII Detection}

Documents are scanned for 8 categories of PII: account numbers, private addresses, private emails, private person names, private phone numbers, private URLs, private dates, and secrets (credentials/tokens/keys). Detected PII text is replaced with category-specific placeholders.

\begin{quote}
\small\ttfamily
Analyze the following text extracted from a document. Determine whether it contains any PII (Personally Identifiable Information).\\[4pt]
Classify each PII finding into one of the following categories:\\
1. account\_number --- Financial or government-issued account identifiers.\\
2. private\_address --- A physical address specific enough to locate a private residence.\\
3. private\_email --- An email address belonging to a specific private individual.\\
4. private\_person --- A real person's name in a private or personal context.\\
5. private\_phone --- A telephone number belonging to a specific individual.\\
6. private\_url --- A URL that exposes personal information.\\
7. private\_date --- A specific date tied to an identifiable individual.\\
8. secret --- Sensitive credentials, tokens, or keys.\\[4pt]
Respond with ONLY a JSON object (no markdown):\\
\{"has\_pii": true/false, "pii\_items": [\{"category": "<category>", "text": "<exact PII substring>"\}, ...]\}\\[4pt]
Text to analyze:
\end{quote}

\subsection{Face Detection in Images}

Document pages are rendered as screenshots and checked for human face photographs.

\begin{quote}
\small\ttfamily
Look at this document page screenshot. Determine whether it contains any human faces (photographs, ID photos, headshots, etc.).\\[4pt]
Do NOT flag:\\
- Clip art, icons, or cartoon illustrations\\
- Logos that happen to contain stylized faces\\
- Very small, unrecognizable figures in diagrams\\[4pt]
Only flag clear, identifiable photographs or images of real human faces.\\[4pt]
Respond with ONLY a JSON object (no markdown):\\
\{"has\_face": true/false\}
\end{quote}

\section{Human Annotation}
\label{sec:human_annotation}

\subsection{Annotators}

We recruit 9 crowdsourced annotators, each of whom is compensated for their work and provides informed consent. The data collection protocol, including human annotation, was approved by the relevant institutional ethics review board. The same 9 annotators perform both data verification and human-baseline editing. Based on annotation results, filtering removed 9 instances for PII, 15 for problematic references, and 12 for unclear instructions (with 4 overlapping between the latter two categories), yielding 32 unique removals and 280 remaining base instances.

\subsection{Task Allocation}

Annotators are divided into 3 groups of 3 (groups A, B, and C), and the 312 candidate base instances are partitioned into 3 balanced subsets of 104 instances. Each annotator completes one instruction variant for every base instance, with 104 tasks under each instruction condition, and never sees multiple instruction variants of the same base instance. The subset-to-condition mapping is rotated across groups, as shown in Table~\ref{tab:human_task_allocation}, so that group effects are balanced across conditions. Because each group contains 3 annotators, every base-instance--condition pair receives 3 independent human outputs. This yields 2,808 candidate annotations and 2,520 annotations after benchmark filtering, with 840 retained annotations per condition. Tasks are presented in a randomized order.

\begin{table}[t]
\centering
\small
\setlength{\tabcolsep}{3pt}
\caption{Counterbalanced assignment of instruction conditions to annotator groups and base-instance subsets.}
\label{tab:human_task_allocation}
\begin{tabular}{lccc}
\toprule
Group & Subset 1 & Subset 2 & Subset 3 \\
\midrule
A & Referential & Descriptive & Inferential \\
B & Descriptive & Inferential & Referential \\
C & Inferential & Referential & Descriptive \\
\bottomrule
\end{tabular}
\end{table}

\subsection{Editing Time}

Annotators face no enforced per-task time limit. Editing time is measured from opening the editable document to declaring the edit complete; time spent on the subsequent evaluation questionnaire is excluded. After benchmark filtering, each instruction condition contains 840 timed annotations, corresponding to 280 base instances completed by 3 annotators. Table~\ref{tab:human_editing_time} reports the resulting time statistics.

\begin{table}[t]
\centering
\small
\caption{Human editing time in seconds by instruction condition. No per-task time limit is enforced.}
\label{tab:human_editing_time}
\begin{tabular}{lrr}
\toprule
Condition & Mean & Median \\
\midrule
Descriptive & 424 & 301 \\
Referential & 362 & 230 \\
Inferential & 324 & 230 \\
\midrule
Overall & 370 & 253 \\
\bottomrule
\end{tabular}
\end{table}

\subsection{Annotation Flow}
\label{sec:full_annotation_instructions}

Each annotation proceeds in two phases: editing and evaluation. We provide the participant-facing editing instructions, questionnaire prompts, and response definitions below. Before starting, annotators see:
\begin{quote}
\small
Please read the annotation guide (README.docx) before starting. Enter your name and select your assignment to begin.
\end{quote}

\paragraph{Phase 1: Editing.}
The interface displays the task's editing instruction and the following directions:
\begin{quote}
\small
Read the instruction below. Click Start Editing to open the document in Word (may take a moment to open). Follow the instruction to modify the document, save the file (Ctrl+S), then return here and click Done Editing.
\end{quote}
Clicking ``Start Editing'' opens a copy of the input document in Microsoft Word and starts the timer. Annotators can use Word functionality such as formatting panels and format painter. They have no access to the reference document during this phase and face no time limit. Clicking ``Done Editing'' ends the editing timer and opens the evaluation phase.

\paragraph{Phase 2: Evaluation.}
The interface displays the task's editing instruction again and provides ``Reopen My Submission'' and ``Open Reference Answer'' buttons. It asks annotators to rate the instruction and input document, compare their edited document with the reference, and report their overall confidence. The participant-facing prompts and response definitions are:
\begin{quote}
\small
Rate the following aspects, then compare your submission with the reference answer.

\textbf{Instruction Clarity.} How clear, understandable, and executable is the editing instruction?

1 --- Very Unclear: Completely incomprehensible or impossible to execute.

2 --- Unclear: Mostly hard to understand; only a few points are executable.

3 --- Acceptable: Somewhat understandable; average clarity.

4 --- Clear: Mostly clear and executable; only minor points are unclear.

5 --- Very Clear: Perfectly clear with no issues in understanding or execution.

\textbf{Task Difficulty.} How much time and effort is needed to complete this task?

Low: Simple and effortless; requires almost no effort.

Medium: Moderate difficulty; requires a fair amount of time and effort.

High: Very challenging; requires significant time and effort.

\textbf{Input Document Quality.} Rate the visual presentation and richness of the input document.

1 --- Very Poor: Almost no substantive content, or extremely messy formatting.

2 --- Poor: Thin content or poor visual presentation.

3 --- Acceptable: Average content and visual presentation.

4 --- Good: Fairly rich content with good visual presentation.

5 --- Very Good: Rich content with excellent visual presentation.

\textbf{Does the document contain PII (personally identifiable information)?} e.g.\ real names, emails, phone numbers, addresses, ID numbers.

Yes: The document contains real PII.

No: The document does not contain any PII.

Unsure: Cannot determine whether certain information is real PII.

\textbf{Is the reference answer correct?} Judge whether the reference itself correctly fulfills the instruction, regardless of your own approach.

1 --- Incorrect: The reference has clear errors and does not correctly follow the instruction.

2 --- Partially Correct: The reference is partly correct but has omissions or errors.

3 --- Correct: The reference correctly fulfills the instruction.

Unsure: Cannot determine correctness (e.g.\ the instruction itself is unclear).

If not fully correct, what is wrong with the reference?

\textbf{Is your result consistent with the reference?} Compare your submission with the reference, regardless of which is more correct.

1 --- Very Different: Your result differs significantly from the reference.

2 --- Partially Similar: Your result partially matches the reference with notable differences.

3 --- Identical: Your result is identical or nearly identical to the reference.

If not fully consistent, what differs?

\textbf{How confident are you in your annotations above?} Your overall confidence in all the evaluations above.

Low: Uncertain about multiple evaluations; may need re-review.

Medium: Confident in most evaluations; uncertain about a few.

High: Confident in all evaluations.
\end{quote}

\section{Common Attribute Errors}
\label{sec:common_target_errors}

We compare attribute errors across all scoreable outputs from the 9 human annotators and 6 agent systems after benchmark filtering. For each output document and attribute, we compute the proportion of its evaluated values that match the reference, and macro-average this accuracy across documents involving that attribute. Table~\ref{tab:common_target_errors} reports the 10 lowest-accuracy attributes for each group. For human outputs, we include attributes evaluated in at least 30 documents. Column width, cell margin, and cell spacing appear among the 10 lowest-accuracy attributes for both groups.

\begin{table*}[t]
\centering
\small
\setlength{\tabcolsep}{4pt}
\caption{Lowest macro-averaged attribute accuracies (\%) for human and agent outputs. $N$ is the number of scoreable output documents in which the attribute is evaluated.}
\label{tab:common_target_errors}
\begin{tabular}{@{}r lrr lrr@{}}
\toprule
& \multicolumn{3}{c}{Human} & \multicolumn{3}{c}{Agent} \\
\cmidrule(lr){2-4} \cmidrule(lr){5-7}
Rank & Attribute & Acc. & $N$ & Attribute & Acc. & $N$ \\
\midrule
1  & Column width             & 20.5 & 99  & Column width             & 36.0 & 174 \\
2  & Object style             & 22.9 & 297 & Cell spacing             & 36.1 & 61 \\
3  & Cell margin              & 28.8 & 171 & Series error bars        & 38.7 & 31 \\
4  & Object alignment         & 37.5 & 72  & Data-table legend key    & 46.4 & 112 \\
5  & Hyperlink                & 40.6 & 429 & Chart data table         & 46.4 & 112 \\
6  & Series line color        & 41.6 & 135 & Soft-edge radius         & 47.1 & 17 \\
7  & Cell vertical alignment  & 43.0 & 153 & Cell margin              & 49.3 & 297 \\
8  & Cell spacing             & 44.4 & 36  & Layout name              & 49.6 & 403 \\
9  & Data-label position      & 45.8 & 54  & Quick style              & 49.7 & 334 \\
10 & Cell shading             & 46.6 & 166 & Minor gridlines          & 50.0 & 32 \\
\bottomrule
\end{tabular}
\end{table*}

\section{Heuristic Baselines for Inferential Variants}
\label{sec:heuristic_baselines}

Inferential variants have lower pass rates than referential variants. The core difference is that inferential variants require agents to identify which object in the document serves as the formatting exemplar, whereas referential variants directly provide this information. To characterize how much of this gap may stem from exemplar localization, we design two heuristic baselines that measure the localization ability of naive strategies.

For each target object, both baselines select a same-type candidate from the same document and check whether its attributes exactly match the ground truth. \textbf{Nearest Neighbor} selects the candidate closest in document order. \textbf{Majority Vote} computes a formatting signature for each same-type candidate, selects the one belonging to the most frequent signature group, and falls back to nearest neighbor when no clear majority exists. Nearest Neighbor achieves 50.6\% and Majority Vote achieves 45.8\%. The superiority of nearest neighbor over majority vote indicates that spatial proximity provides a stronger signal than global format consensus---same-type objects in a document do not always share the same formatting. For example, when the target is a heading paragraph, the majority format among same-type candidates likely corresponds to body text rather than headings, causing majority vote to select incorrectly.

Note that these heuristic baselines only evaluate reference object localization accuracy (whether the selected candidate's attributes match the ground truth), without performing actual document editing. They are therefore not directly comparable to end-to-end agent pass rates, which additionally require correct execution of attribute modifications.

\section{Visual Attribute Detection Experiment}
\label{sec:visual_example_appendix}

For each of the 280 document pairs, we first identify the pages where the input and reference documents differ and render both versions as screenshots, retaining at most three pages per document. GPT-5.5 receives these screenshots together with an 81-attribute checklist and selects every attribute that differs visually. We evaluate its selections against ground-truth attribute diffs computed programmatically from document properties. Because the model can directly compare the input with the reference and choose from a fixed checklist, this controlled setup narrows the judgment space and is easier than the actual editing scenario.

\begin{figure*}[t]
\centering
\begin{minipage}[c]{0.48\textwidth}
\docframe{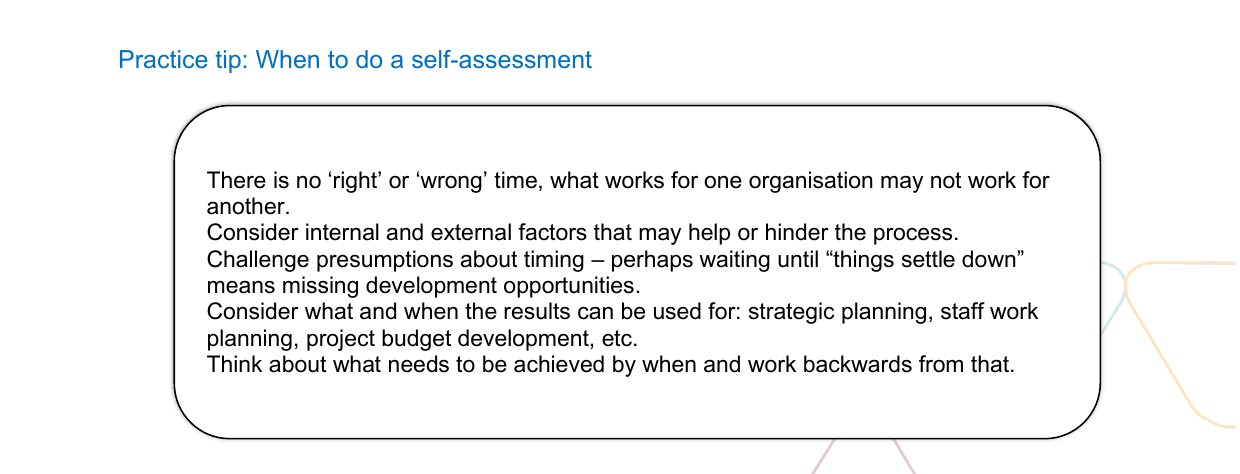}{Input document}

\vspace{8pt}

\docframe{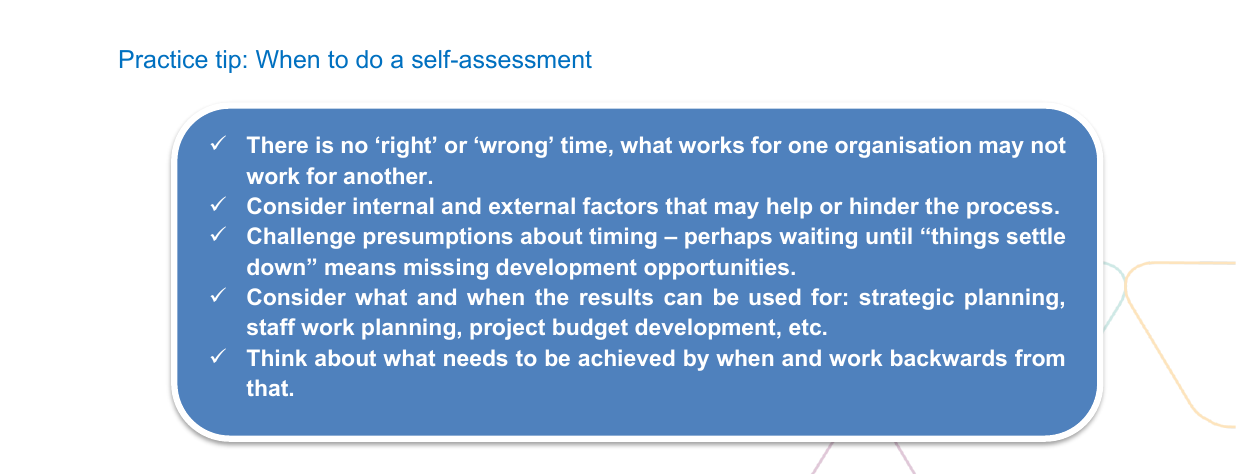}{Reference document}
\end{minipage}
\hspace{16pt}
\begin{minipage}[c]{0.30\textwidth}
\begin{tikzpicture}
\node[inner sep=0pt, anchor=north west] (card) at (0,0) {%
\begin{minipage}{\dimexpr\linewidth-1pt\relax}%
\vspace{14pt}%
\par\vspace{6pt}%
\centering
\begin{tabular}{@{}c@{\hspace{4pt}}l@{}}
{\color{green!60!black}\cmark} & \small Fill color\\[3pt]
{\color{green!60!black}\cmark} & \small Bold\\[3pt]
{\color{green!60!black}\cmark} & \small Line weight\\[3pt]
{\color{green!60!black}\cmark} & \small List type\\[3pt]
& \small\ldots
\end{tabular}\vspace{6pt}%
\end{minipage}};
\draw[gray!30, line width=0.3pt, rounded corners=3pt] (card.south west) rectangle (card.north east);
\fill[green!60!black, rounded corners=3pt] (card.north west) rectangle ([yshift=-14pt]card.north east);
\fill[green!60!black] ([yshift=-7pt]card.north west) rectangle ([yshift=-14pt]card.north east);
\node[font=\small\sffamily\bfseries, text=white] at ([yshift=-7pt]card.north) {Detected};
\end{tikzpicture}

\vspace{12pt}

\begin{tikzpicture}
\node[inner sep=0pt, anchor=north west] (card) at (0,0) {%
\begin{minipage}{\dimexpr\linewidth-1pt\relax}%
\vspace{14pt}%
\par\vspace{6pt}%
\centering
\begin{tabular}{@{}c@{\hspace{4pt}}l@{}}
{\color{red!70!black}\xmark} & \small Line spacing\\[3pt]
{\color{red!70!black}\xmark} & \small Spacing after\\[3pt]
{\color{red!70!black}\xmark} & \small Shadow angle\\[3pt]
{\color{red!70!black}\xmark} & \small Shadow distance\\[3pt]
& \small\ldots
\end{tabular}\vspace{6pt}%
\end{minipage}};
\draw[gray!30, line width=0.3pt, rounded corners=3pt] (card.south west) rectangle (card.north east);
\fill[red!70!black, rounded corners=3pt] (card.north west) rectangle ([yshift=-14pt]card.north east);
\fill[red!70!black] ([yshift=-7pt]card.north west) rectangle ([yshift=-14pt]card.north east);
\node[font=\small\sffamily\bfseries, text=white] at ([yshift=-7pt]card.north) {Missed};
\end{tikzpicture}
\end{minipage}
\caption{Example from the visual attribute detection experiment. GPT-5.5 correctly identifies visually salient changes such as fill color and bold weight, but misses subtle attributes like line spacing, spacing after, and shadow parameters whose changes produce barely perceptible visual differences.}
\label{fig:visual_example}
\end{figure*}

The example in Figure~\ref{fig:visual_example} reflects the aggregate precision--recall pattern. GPT-5.5 usually identifies visually salient changes such as color and bold weight, but often overlooks subtler differences in line spacing, paragraph spacing, font, and shadow parameters. Its high precision indicates that detected changes are generally correct, whereas its low recall shows that screenshot inspection recovers only about one-third of the actual changes. Because coding agents may also inspect document properties programmatically through COM or XML, these results should be interpreted as evidence of the limitations of screenshot-only attribute detection and vision-only LLM judges, rather than as a complete explanation of agent failures.

\section{Preservation Results}
\label{sec:preservation_results}

Table~\ref{tab:preservation} reports preservation pass rates for each model and instruction condition. A variant passes only when no differences are detected in any evaluated non-target object.

\begin{table}[t]
\centering
\small
\setlength{\tabcolsep}{3pt}
\caption{Preservation pass rates (\%) across the three instruction conditions, using all 280 tasks per condition as the denominator. The average is computed over all 840 tasks across the three conditions.}
\label{tab:preservation}
\begin{tabular}{@{}lrrrr@{}}
\toprule
Agent & Desc. & Ref. & Infer. & Avg. \\
\midrule
Claude Code (Opus 4.6) & 92.1 & 90.0 & 89.6 & 90.6 \\
Claude Code (Sonnet 4.6) & 92.1 & 88.6 & 86.4 & 89.0 \\
Codex (GPT-5.5) & 65.4 & 52.9 & 70.0 & 62.7 \\
Claude Code (GLM 5.2) & 91.8 & 86.4 & 92.5 & 90.2 \\
Kimi CLI (Kimi K2.6) & 84.6 & 85.4 & 85.4 & 85.1 \\
Codex (GPT-5.4) & 54.6 & 51.8 & 60.7 & 55.7 \\
\bottomrule
\end{tabular}
\end{table}

\end{document}